\documentclass[runningheads]{llncs}

\usepackage{eccv}

\usepackage{xcolor}
\definecolor{dpurple}{RGB}{102,0,153}

\usepackage{eccvabbrv}

\usepackage{graphicx}
\usepackage{booktabs}
\usepackage{multirow}
\usepackage{wrapfig}
\usepackage[accsupp]{axessibility}  

\usepackage{hyperref}

\usepackage{orcidlink}

\begin{document}

\title{SiPhy: Single-Image Physical Property Reasoning} 


\author{Hoang Le\inst{1}\and
Joonwoo Kwon\inst{1} \and
Elkhan Ismayilzada\inst{1} \and Yufei Zhang\inst{2} \and Zijun Cui\inst{1}\orcidlink{0000-0002-4362-197X} }

\authorrunning{H. Le et al.}

\institute{Michigan State University, East Lansing MI 48864, USA \and
Independent Researcher}

\maketitle

\renewcommand{\labelitemi}{\textbullet}
\newcommand{\methodname}{SiPhy\xspace}
\newcommand{\myhat}[1]{\small{\color{dpurple}{}
{$_{#1}$}}}
\newcommand{\old}[1]{\textcolor{orange}{#1}}
\newcommand{\hl}[1]{\textcolor{blue}{#1}}
\colorlet{darkgreen}{green!50!black}
\newcommand{\new}[1]{\textcolor{darkgreen}{#1}}

\begin{abstract}
Inferring physical properties such as mass, stiffness, and elasticity from a 
single image is essential for simulation and embodied AI, yet most existing 
approaches rely on multi-view reconstruction or physics-based supervision. 
We introduce \methodname, a unified framework for single-image physical 
property reasoning that aligns 3D-aware visual cues, depth with language-based 
material knowledge. From one RGB image, \methodname samples pseudo-voxel 
points, extracts CLIP features, and grounds them to material candidates 
proposed by an VLM. A part-based contrastive aggregator enforces region 
consistency, while a heaviness-aware refinement improves thickness and volume 
estimation for dense objects. 

Across ABO-500, MVImgNet-100, and PhysXNet-100, \methodname achieves 
state-of-the-art \emph{single-image} performance, surpassing multi-view 
reconstruction methods by improving mass MnRE by up to \textbf{93\%} (vs.\ PUGS), 
reducing density MAE by \textbf{35.5\%} (vs.\ NeRF2Physics), and lowering Young’s modulus 
error by \textbf{23.5\%}. We further validate \methodname on real 
hand–object interaction datasets, demonstrating its potential as a data annotation engine for physical understanding from single-view imagery. Codes are available at \url{https://github.com/DominoAI-Lab/SiPhy-ECCV-2026}. 
\end{abstract}

\section{Introduction}
\label{sec:intro}

Humans can effortlessly infer how heavy, rigid, or flexible an object is from a single glance. A metal mug and a foam cup immediately evoke distinct expectations about mass, stiffness, and density even without motion cues or interaction. This ability to perform single-view physical reasoning plays a crucial role in human perception, supporting behaviors such as grasping, tool use, and anticipating object dynamics \cite{hamrick2011internal}. This naturally raises the question: \textit{Can an AI system learn to infer an object's physical properties from just a single image?}


Despite its importance, this problem remains largely unsolved. Unlike visual attributes, physical properties are latent and cannot be directly observed. They must be inferred from subtle cues in appearance, 3D geometry, and material semantics. Existing methods often rely on multi-view cues~\cite{zhai2024physical,xu2024gaussianproperty} or accurate 3D representation from multi-view image captures (\eg, NeRF~\cite{mildenhall2020nerf}). These approaches achieve strong performance but require many input views and heavy optimization. As a result, they are impractical in everyday scenarios where only a single image is available. On the other hand, single-image physical reasoning is fundamental to many downstream applications including sound synthesis~\cite{dou2025hearing}, virtual editing~\cite{hsu2025autovfx}, simulation~\cite{dagli2025vomp}, and embodied reasoning~\cite{jiang2025phystwin}.

Prior single-image efforts face two main limitations: 1) Lack of 3D geometric awareness. Most prior works focus on pixel-level material recognition~\cite{ Bell2015MaterialRecognitionInTheWild, Sharan2013RecognizingMaterials}, treating the problem purely as a 2D classification task agnostic to the object’s geometry. This prevents them from estimating 3D-consistent physical quantities such as volume or mass. 
2) Lack of physical grounding beyond appearance. Methods such as Image2Mass \cite{pmlr-v78-standley17a} directly regress physical quantities from RGB appearance through data-driven learning, but visual appearance alone provides insufficient clues regarding physical behavior. Without explicit material knowledge or physical knowledge, such models fail to generalize to unseen environments, materials, and object compositions. 

To address these challenges, we propose \textbf{SiPhy}, a \textbf{S}ingle-\textbf{i}mage \textbf{Phy}sical property reasoning framework that unifies 3D geometry, semantics, and language-based physical knowledge. Our key insight is that core advantages of multi-view physical reasoning, including structured geometry, consistent material inference, and physics-aware aggregation, can be approximated from a single RGB image through deliberate architectural design. Siphy consists of three main components: 1) \textbf{3D-Aware Visual Sampling}. We perform geometry-aware 2D sampling that approximates voxel centers when lifted into 3D. This enables coarse but spatially structured reasoning from a single view. 2) \textbf{Material Probability Estimation}. For each sampled region (centered with sampled points), CLIP provides visual embeddings while a fine-tuned Vision Language Model (VLM) proposes material candidates and associated physical attributes such as density, Young's modulus, and thickness. A mask-aware contrastive module aligns visual patches with material semantics and produces material likelihoods for each pseudo-voxel. 3) \textbf{Physical Property Estimation and Refinement}. The physical property value at each point is obtained as the expectation over material likelihoods, and 3D-level quantities (\eg, total mass) are aggregated over all voxels. A heaviness-aware thickness refinement module further improves accuracy as it is directly correlated and sensitive to mass prediction.

Through this design, SiPhy enables both pixel-level and object-level reasoning from a single image. Across diverse dataset including ABO500 \cite{DBLP:journals/corr/abs-2110-06199}, MVImgNet \cite{yu2023mvimgnet}, and PhysXNet100 \cite{cao2025physx}, SiPhy achieves state-of-the-art performance in mass prediction, material segmentation, density estimation, and Young's modulus against both single-view and multi-view baselines. We further demonstrate strong generalization on real hand-object interaction datasets, highlighting its potential for real-world applications. Our contributions are summarized as follows: 

\begin{itemize}
    \item We introduce \textbf{SiPhy}, \textbf{the first single-image and depth framework} capable of predicting both 2D- and 3D-level physical properties.

    \item To unify geometry, semantics, and physical knowledge from a single RGB image, we propose a \textbf{3D-aware vision--language physical reasoning pipeline} that integrates CLIP-based visual grounding, VLM-driven material and attribute inference, and geometry-aware voxelization.

    \item Through comprehensive comparisons, SiPhy achieves state-of-the-art results on ABO500, MVImgNet100, and PhysXNet100 for mass prediction, material segmentation, density estimation, and Young’s modulus. 
    We further validate its generalization on real hand-object interaction datasets and demonstrate its downstream utility by using SiPhy’s material predictions as physical priors for image-to-audio generation.
\end{itemize}

\section{Related Work}
\label{sec:related}

We begin by discussing \textbf{Visual Physical Property Reasoning}, which includes both video-based and image-based approaches. Our work falls into the image-based category. We then broaden our discussion to recent advancement in foundations models' physics reasoning, primarily \textbf{LLMs’ Physics Reasoning}. This involves understanding intuitive physics, which studies how agents or models reason about physical dynamics and object interactions in simulated environments, and text based LLM understanding, where large language models infer or describe physical properties and relationships purely through language.

\textbf{Visual Physical Property Reasoning.}
Reasoning about physical and material properties from visual observations has long been a core challenge in computer vision. Existing research can broadly be divided into two categories. The first performs reasoning given videos where methods such as ~\cite{NIPS2015_d09bf415, NIPS2017_4c56ff4c, xu2019densephysnetlearningdensephysical} infer dynamic physical properties that are closely tied to motion (\eg, mass, friction, ) from videos by coupling visual perception with a physics engine or differentiable dynamics simulator. Therefore, these approaches remain restricted to controlled laboratory settings and are difficult to generalize or apply in real-world scenarios.

The second line of work aims to infer physical properties from static images only (our work is categorized into this line), using either multi-view or single-view setups. Early works tend to work with single-view image, predicting material of objects or real-world scenes~\cite{ Bell2015MaterialRecognitionInTheWild, Sharan2013RecognizingMaterials}. They used CNN or hand-crafted image features. These works often focused on pixel-level property without understanding spatial structure of the object. Due to this, they cannot predict object-level properties such as mass. To enable mass prediction given a single image, image2mass\cite{pmlr-v78-standley17a} introduced the first such dataset ABO containing 500 objects. A Xception-style framework is introduced to predict mass. However, its data-driven nature limits the generalization of the proposed model. 

Recent works advance image-based physical property reasoning through multi-view 3D reconstruction techniques and large foundation models. Multi-view 3D reconstruction enables detailed geometric representations, while large Visual-Language Models (VLMs)~\cite{li2023blip2bootstrappinglanguageimagepretraining, instructblip} and image–text alignment frameworks such as CLIP~\cite{Cherti_2023, khosla2021supervisedcontrastivelearning} enhance semantic reasoning and generalization. For example, NeRF2Physics~\cite{zhai2024physical} leverages NeRF to represent 3D scenes and injects CLIP features to model physical properties at each spatial location. It virtually voxelizes objects by assigning appropriate thickness and size to the reconstructed 3D points, through which object-level properties such as mass can be estimated. Similarly, GaussianProperty~\cite{xu2024gaussianproperty} and PUGS~\cite{shuai2025pugszeroshotphysicalunderstanding} adopt 3D Gaussian Splatting~\cite{kerbl20233dgaussiansplattingrealtime}, where objects are naturally voxelized through Gaussian primitives for efficient physical reasoning. Noticeable limitations of these methods in real-world applications are their tedious and time-consuming process, as well as their reliance on multi-view images, which are often unavailable or costly to obtain.
In this work, we unifies visual-language material reasoning and 3D-aware physical inference and propose a single-image physical property reasoning model, SiPhy, that can infer both pixel-level and object-level properties.

\textbf{LLMs' Physics Reasoning.} 
Majority of the efforts focus on proposing benchmarks for evaluating LLM's physics reasoning through visual question answering~\cite{bakhtin2019phyrenewbenchmarkphysical, li2024iphyreinteractivephysicalreasoning, zheng2024contphycontinuumphysicalconcept, cherian2024llmphycomplexphysicalreasoning, chen2022comphycompositionalphysicalreasoning, bordes2025intphys2benchmarkingintuitive}. These works test agent physic understanding through quiz performed in 2D simulator. These are mainly video-based, covering intuitive physics in the form of quiz and yes/no questions.

Previous works also explored adopting large language models (LLMs) to predict physical and material properties directly from textual or other symbolic descriptions such as chemical compositions or crystal structures \cite{chaudhari2024alloybertalloypropertyprediction, LI2024376, jacobs2024regressionlargelanguagemodels, rubungo2023llmproppredictingphysicalelectronic}. These works demonstrated that LLMs can regress quantities like material, density, band gap, or elastic modulus, though they remain limited to text inputs. SiPhy follows and expands these works by enabling  properties prediction with multi-modality input such as image, text and functional descriptions of objects. 

\begin{figure*}[h!]
  \centering

  \includegraphics[width=\linewidth]{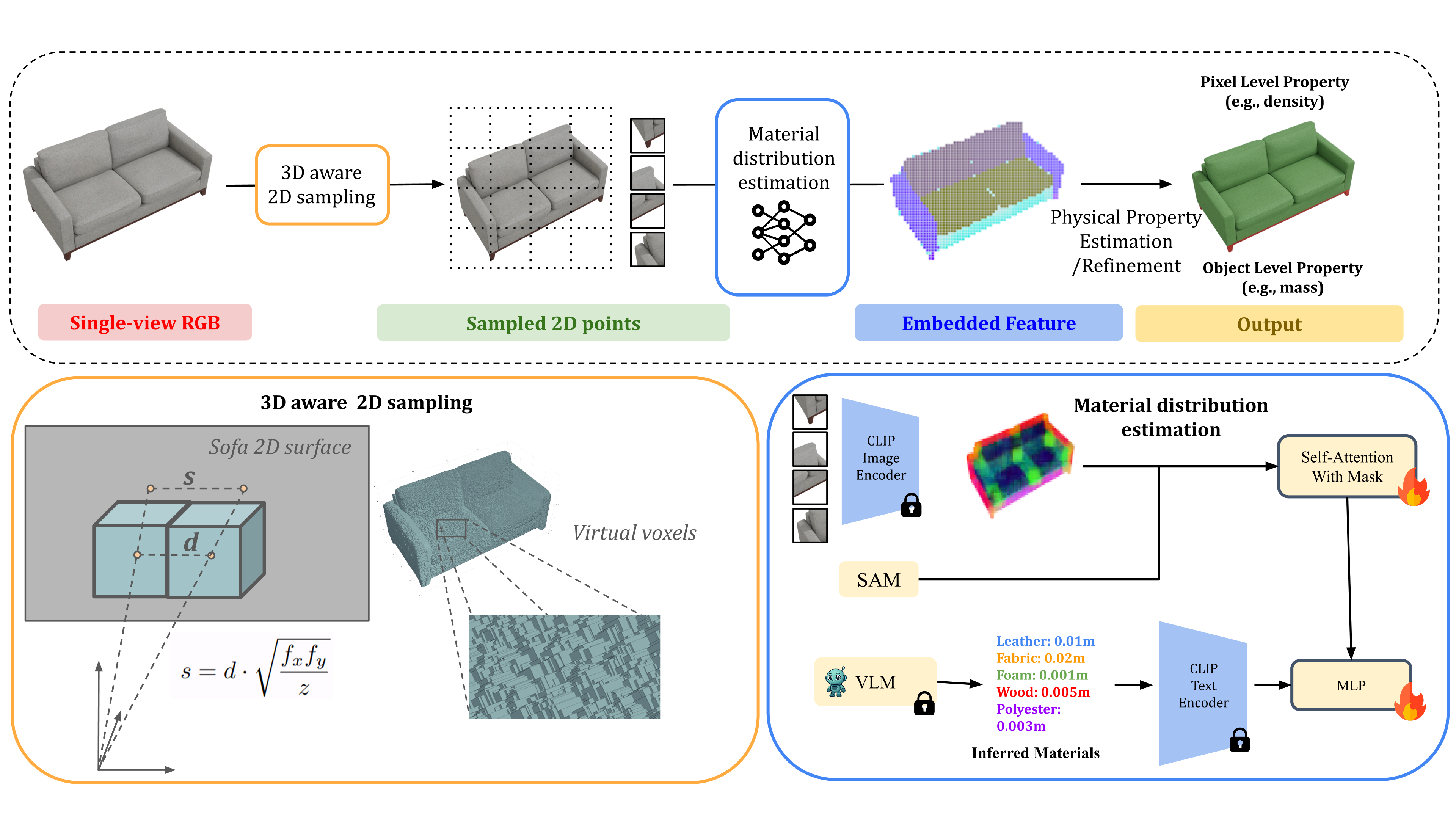}
  \caption{\textbf{Overall Architecture of the SiPhy.} 
(1) \textbf{3D-aware 2D sampling:} spatially distributed 2D points are sampled from the object surface to approximate coarse geometry from a single view. 
(2) \textbf{Material probability estimation:} CLIP visual features, guided by SAM masks, are aligned with material candidates proposed by a finetuned VLM to produce material probability maps. 
(3) \textbf{Physical property prediction:} localized material probabilities are integrated over the sampled points to infer the properties
}

\label{fig:overall_pipeline}

\end{figure*}

\section{Method}

Given a single RGB image $I\in\mathbb{R}^{H\times W\times3}$, our goal is to infer both pixel-level physical property maps, such as density or Young's modulus, and global object-level quantities like total mass.

We first introduce the SiPhy VLM, a separately fine-tuned vision--language model that proposes material candidates used by the subsequent physical reasoning pipeline (Sec.~\ref{subsec: VLM}). As shown in~\cref{fig:overall_pipeline}, the pipeline consists of three interconnected stages. First, we perform 3D-aware visual sampling to identify spatially distributed 2D points on the object surface, which approximate a coarse geometric structure from a single view (Sec.~\ref{subsec:sampling}). Second, CLIP encodes the patch around each sampled point, and the resulting visual features are aligned with VLM-proposed material candidates to estimate point-wise material probabilities (Sec.~\ref{subsec:material_prob}).
Finally, we compute the physical properties by integrating localized material likelihoods across the sampled points and apply a heaviness-aware refinement module to stabilize the final object-level predictions (Sec.~\ref{subsec:predict}).



\subsection{SiPhy VLM}
\label{subsec: VLM}
To infer part-level materials, SiPhy uses a fine-tuned VLM. For each SAM-derived object part, the VLM predicts $K$ candidate materials, which are encoded by the CLIP text encoder and used by the material probability estimation stage (Sec.~\ref{subsec:material_prob}).

\begin{figure*}[ht!]
  \centering
  \includegraphics[width=\linewidth]{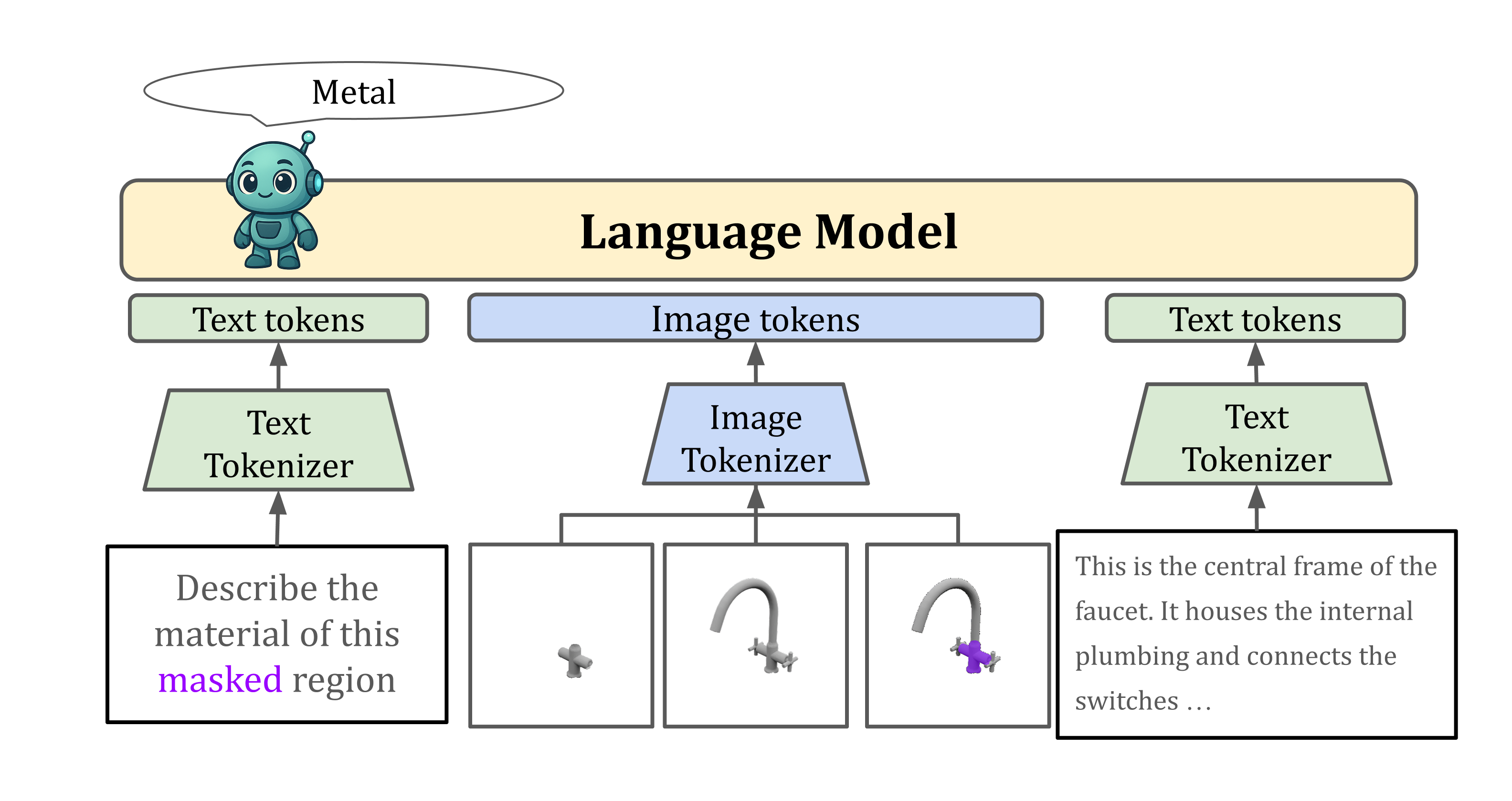}
  \caption{\textbf{SiPhy VLM} (Sec.~\ref{subsec: VLM}). For each SAM-derived object part, the model receives three visual inputs (part mask, cropped region, full object image) alongside a text prompt and a GPT-4-generated part description. Vicuna-7B-v1.5 jointly processes all inputs and predicts the material category (e.g., ``Metal'').}
  \label{fig:siphy_vlm}
\end{figure*}

The VLM follows a standard LVLM design~\cite{liu2023visualinstructiontuning, yu2024octopi, alayrac2022flamingovisuallanguagemodel} with a frozen CLIP ViT-L/14 visual encoder~\cite{ilharco_gabriel_2021_5143773}, a two-layer MLP projector, and Vicuna-7B-v1.5~\cite{vicuna2023} as the language backbone. For each object part, it takes three visual inputs: the part mask, the cropped part image, and the full object image (Fig.~\ref{fig:siphy_vlm}). The three CLS features are assigned positional encodings, projected into Vicuna's token space, and concatenated with the text prompt and GPT-4-generated part description for material prediction.

Training uses two stages. We first train only the projection module on approximately 10k PhysX- and MVImgNet-derived samples, while keeping the visual encoder and LLM frozen. We then fine-tune Vicuna with LoRA and jointly train the projector, with the visual encoder still frozen. We treat material prediction as a classification task: the attention-masked mean-pooled last hidden state of the LLM is passed through a linear classifier and optimized with cross-entropy loss, reducing the output space to $K$ material classes rather than the full LLM vocabulary. At inference, the predicted material names are encoded by CLIP and passed to Sec.~\ref{subsec:material_prob}.

\subsection{3D-Aware Visual Sampling}
 \label{subsec:sampling}

Pixel-level properties can be directly inferred from local visual cues and are closely related to semantic segmentation tasks. In contrast, object-level property prediction, such as total mass, requires a thorough spatial understanding of the object. Such an understanding necessitates 3D voxelization, which involves partitioning the object using a \textit{unit object} representation, such as virtual cubes or Gaussians.

Existing methods adopt two primary voxelization strategies. NeRF2Physics~\cite{zhai2024physical} utilizes virtual cubes. We refer to these as \textit{virtual} because the unit cubes do not explicitly exist; rather, they are represented implicitly by their centers' 2D projections and their 3D dimensions. Alternatively, GaussianProperty~\cite{xu2024gaussianproperty} and PUGS~\cite{shuai2025pugszeroshotphysicalunderstanding} reconstruct objects using Gaussian Splatting, where Gaussian primitives naturally serve as voxels. In this work, we follow the virtual cube voxelization approach, where each unit is parameterized by its edge length $d$ and its thickness. Notably, thickness is not directly observed but is inferred from the predicted material distribution, for instance, by computing its expectation under the material probability field.

\begin{figure}[h]
\vspace{-6pt}
\centering
\begin{subfigure}{0.45\columnwidth}
    \centering
    \includegraphics[width=\linewidth,height=0.80\linewidth]{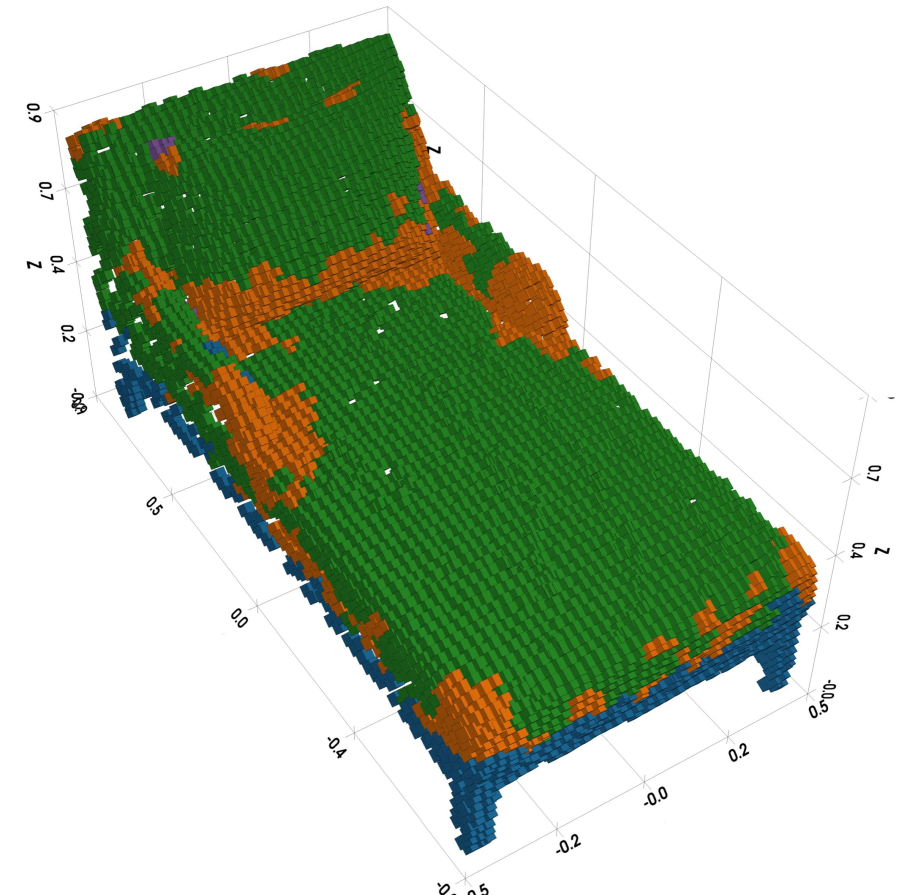}
    \caption{Virtual voxelization}
\end{subfigure}
\hfill
\begin{subfigure}{0.45\columnwidth}
    \centering
    \includegraphics[width=\linewidth,height=0.65\linewidth]{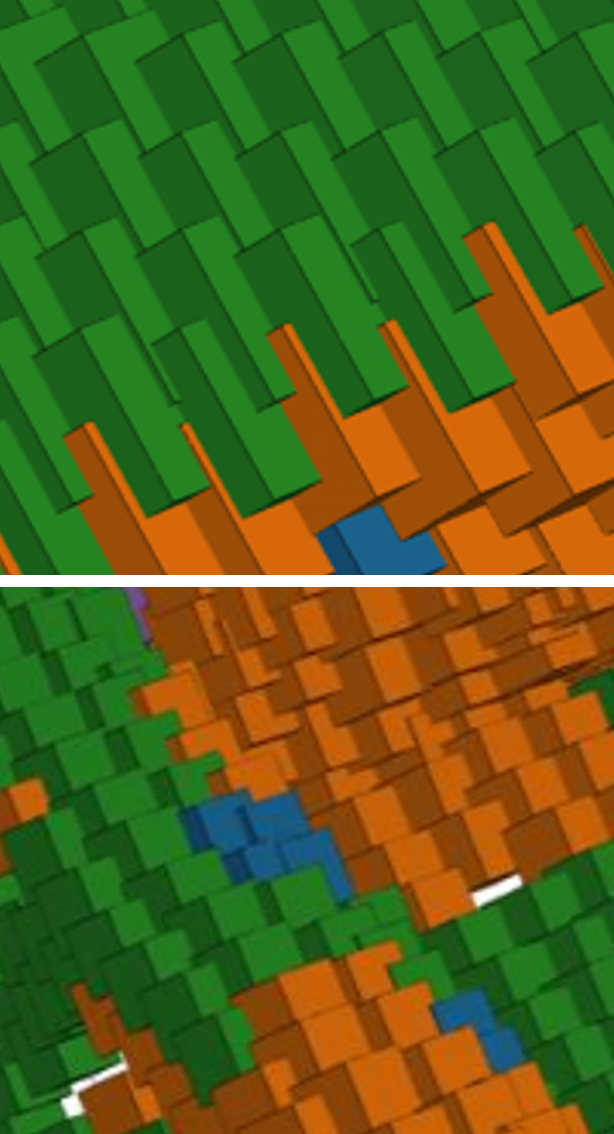}
    \caption{Zoomed. \textbf{Up:} Uniform overlapping voxels, \textbf{Down:} Ours, not overlapping voxels}
\end{subfigure}
\caption{\textbf{3D-aware pseudo-voxelization.} (a) Depth-adaptive sampling voxels (colored rectangulars). (b) Zoomed comparison: fixed uniform spacing (top) causes voxel overlap and redundancy; our adaptive spacing (bottom) produces non-overlapping coverage that better approximates a surface voxel grid.}
\label{fig:virtual_voxelization}
\vspace{-6pt}
\end{figure}

Our objective is to derive the 2D projections of these virtual 3D unit cubes, conditioned on a single RGB image and a predefined side length $d$ (side length of the virtual unit cube). While previous efforts like NeRF2Physics~\cite{zhai2024physical} necessitate multi-view reconstruction to facilitate 3D point downsampling and projection, we demonstrate that a structured representation is attainable using only single-view imagery and metric depth estimates. To maintain consistent spatial coverage and approximate a pseudo-voxel grid from a single perspective, we determine the adaptive 2D pixel spacing $s$ as follows:
\begin{equation}
s = d \cdot \sqrt{\frac{f_x f_y}{z}}
\end{equation}
where $f_x$ and $f_y$ denote the camera intrinsics, and $z$ represents the estimated object depth. In scenarios characterized by significant background clutter, we employ a monocular depth model (e.g., Depth Anything~\cite{depth_anything_v2}) to provide a stable geometric prior. Fig.~\ref{fig:overall_pipeline} illustrates how this sampling strategy yields a physically grounded distribution of 2D unit cube projections. Our experiments confirm that a naive fixed-spacing baseline results in suboptimal performance (\cref{tab:mass}), primarily due to virtual voxel overlap and redundancy (\cref{fig:virtual_voxelization}). Since $s$ depends on estimated depth and camera intrinsics, we analyze sensitivity to perturbations of these quantities in the Supplementary.

Using the spacing $s$, we sample $N$ non-overlapping 2D points over the object. A square patch of side length $s$ is cropped around each point and embedded using a frozen CLIP ViT-B/16 encoder~\cite{ilharco_gabriel_2021_5143773}, yielding visual descriptors $\{f_i\}_{i=1}^N \in \mathbb{R}^D$. This stage provides a spatially structured set of
3D-aware visual tokens that approximate voxel centers for subsequent reasoning. The detailed sampling algorithm is presented in Supplementary materials.

\subsection{Material Probability Estimation}
\label{subsec:material_prob}
To infer material categories and their associated physical attributes from local appearance cues, SiPhy integrates CLIP-based visual features, VLM-driven material reasoning (Sec.~\ref{subsec: VLM}), and mask-aware contrastive alignment. This stage produces a material likelihood matrix $P \in \mathbb{R}^{N \times K}$ for the $N$ sampled points across $K$ candidate materials.

\noindent\textbf{CLIP visual and text embeddings.}
Motivated by the sophisticated world knowledge embedded in vision–language models like LLaVA~\cite{liu2023visualinstructiontuning}, we leverage the fine-tuned VLM (Sec.~\ref{subsec: VLM}) to propose candidate materials and their intrinsic physical properties (\eg, density, Young's modulus, and thickness) in textual form.


As mentioned above, each sampled patch is encoded via a frozen CLIP ViT-B/16 encoder~\cite{ilharco_gabriel_2021_5143773} to extract visual features $\{\mathbf{f}_i\}_{i=1}^{N}$. Simultaneously, the candidate material names generated by the VLM are projected into the same latent space using the CLIP text encoder, resulting in embeddings $\{\mathbf{t}_k\}_{k=1}^{K}$. While the cosine similarity between these modalities provides an initial estimation, these likelihoods are subsequently refined through a spatially-aware alignment process to ensure physical consistency. We denote the resulting raw similarity matrix as $S_{\text{CLIP}} \in \mathbb{R}^{N \times K}$, where $[S_{\text{CLIP}}]_{i,k} = \mathrm{sim}(\mathbf{f}_i, \mathbf{t}_k)$.


\noindent\textbf{Part-based contrastive alignment.}
Material is generally consistent within the same physical part of an object. 
However, a direct CLIP similarity between visual and text embeddings does not explicitly enforce this structure. 
To incorporate \textit{part-level coherence}, we introduce a part-based contrastive alignment module that encourages points within the same part to share similar material predictions, while separating points belonging to different parts.

For each sampled feature $\mathbf{f}_i$, we define the part-aware neighborhood $\mathcal{N}(i)$ as the set of points that fall within the same SAM-derived part mask as $i$. 
We then apply self-attention restricted to this part region:
\begin{equation}
\tilde{\mathbf{f}}_i =
\mathrm{softmax}\!\left(
    \frac{(\mathbf{f}_i W_Q)
    (\mathbf{F}_{\mathcal{N}(i)} W_K)^\top}
    {\sqrt{d_a}}
\right)
\mathbf{F}_{\mathcal{N}(i)} W_V ,
\end{equation}
where $\mathbf{F}_{\mathcal{N}(i)} \in \mathbb{R}^{|\mathcal{N}(i)| \times D}$ is the matrix of CLIP features of all sampled points in the same SAM-derived part as $i$, $W_Q, W_K \in \mathbb{R}^{D \times d_a}$ and
$W_V \in \mathbb{R}^{D \times D}$ are learnable projections and $d_a$ is the attention dimension.

The part-refined features are then concatenated with the corresponding material text embeddings and passed through a MLP, producing output $\mathbf{z}_i \in \mathbb{R}^D$ for each sampled point $i$.

We optimize the similarity matrix $S$ using a supervised contrastive loss~\cite{khosla2021supervisedcontrastivelearning} defined over part groupings:
\begin{equation}
\mathcal{L}_{\text{CL}} =
-\frac{1}{N}\sum_i
\frac{1}{|\mathcal{P}(i)|}
\sum_{p \in \mathcal{P}(i)}
\log
\frac{\exp(\mathrm{sim}(\mathbf{z}_i, \mathbf{z}_p)/\tau)}
{\sum_{a \neq i}\exp(\mathrm{sim}(\mathbf{z}_i, \mathbf{z}_a)/\tau)},
\end{equation}
where $\mathcal{P}(i)$ denotes the set of samples belonging to the same SAM-derived part as $i$, $\tau$ is a temperature scalar, and $\mathrm{sim}(\mathbf{u}, \mathbf{v})=\mathbf{u}^\top \mathbf{v}$ is the cosine similarity between embeddings.

We use $S_{\text{CLIP}}$ as a regularizer to prevent the module from smoothing embeddings without preserving the material discrimination for each 2D point:
\begin{equation}
\mathcal{L}_{\text{align}} =
\frac{1}{NK}\,
\mathcal{D}\!\left(S_{\text{MLP}},\, S_{\text{CLIP}}\right),
\end{equation}
where $\mathcal{D}(\cdot,\cdot)$ denotes the distance between the learned and teacher similarity matrices, instantiated as either
$\ell_2$ distance $\|S_{\text{MLP}} - S_{\text{CLIP}}\|_F^2$
or cross-entropy divergence. The overall training objective is:
\begin{equation}
\mathcal{L}_{\text{total}} =
\mathcal{L}_{\text{CL}} +
\lambda\,\mathcal{L}_{\text{align}},
\end{equation}
where $\lambda$ balances the alignment strength.
This joint formulation encourages intra-part smoothness while keeping the learned similarities consistent with CLIP’s embedding geometry.

We also tried a naive approach of averaging the visual features of all points in 1 mask, which does not perform well, as shown in~\cref{tab:mass}.

\subsection{Physical Property Estimation and Refinement}
\label{subsec:predict}
\noindent\textbf{Pixel-level property prediction.} Given the material likelihoods produced by the vision-language reasoning stage, we compute the physical property at each sampled point as the expectation over material-specific attributes.

Let $P \in \mathbb{R}^{N \times k}$ denote the likelihood matrix for $N$ sampled
points and $K$ candidate materials. 
The predicted property at point $i$ is:
\begin{equation}
    \hat{V}_i = \sum_{j=1}^{k} p_{i,j} V_j,
\end{equation}
where $p_{i,j}$ is the likelihood of material $j$ and $V_j$ is its corresponding
attribute (e.g., density or thickness). The probabilities are normalized such that $\sum_{j=1}^{k} p_{i,j} = 1$ for each $i$. Values for non-sampled pixels are propagated through $k$-Nearest Neighbor interpolation.

\noindent\textbf{Object-level property prediction.} To calculate object level property, such as mass, we follow NeRF2Physics~\cite{zhai2024physical} to  treat thickness as a property like mass density, and we calculate the expected volume and mass of each point (as if there is a voxel around that point). Eventually, the mass of the object is obtained as the sum of the mass of each point.  

\begin{figure}[tb]
  \centering
  \begin{subfigure}{0.48\linewidth}

    \includegraphics[width=1\linewidth]{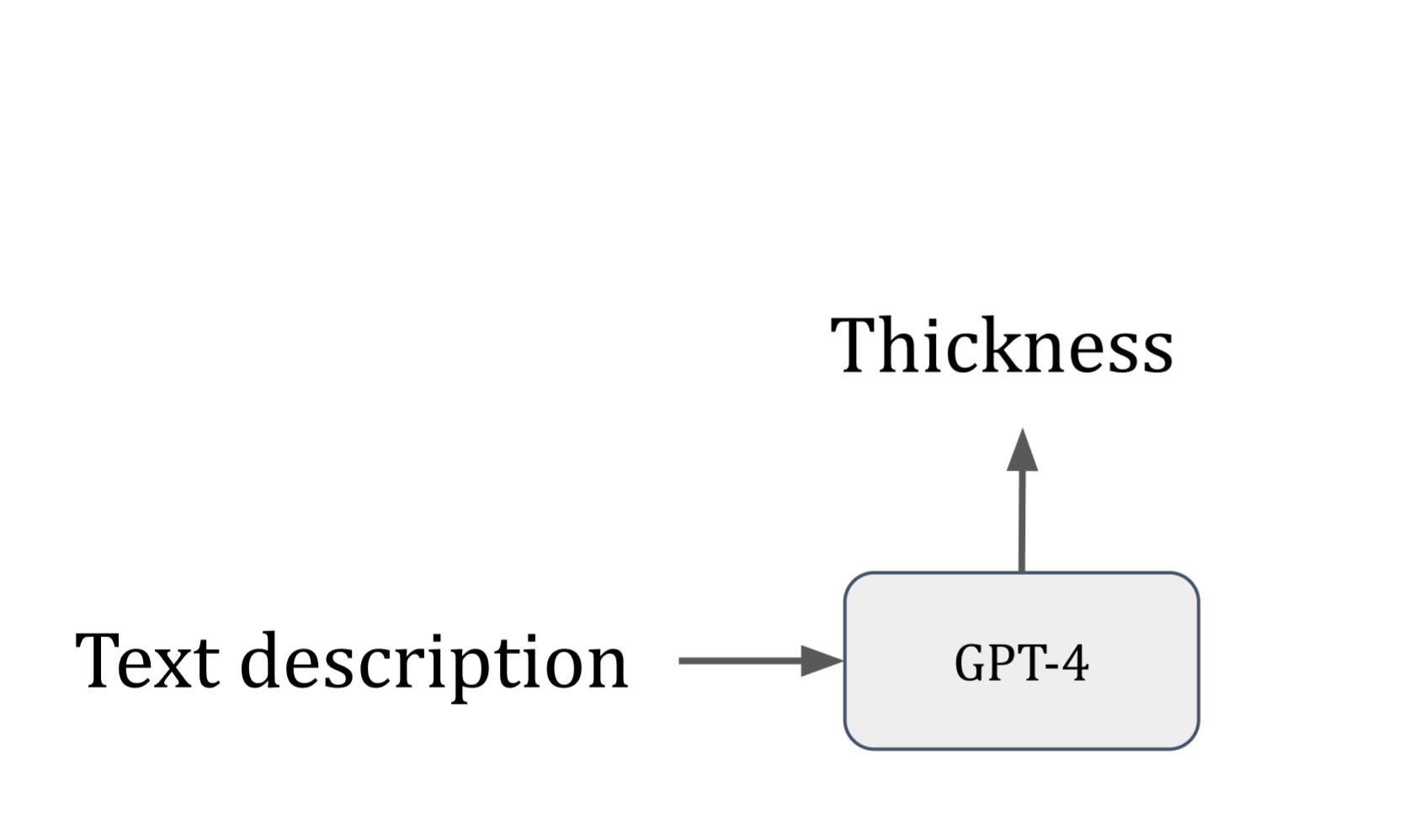}
    \caption{NeRF2Physics}
    \label{fig:short-a}
  \end{subfigure}
  \hfill
  \begin{subfigure}{0.48\linewidth}


    \includegraphics[width=1\linewidth]{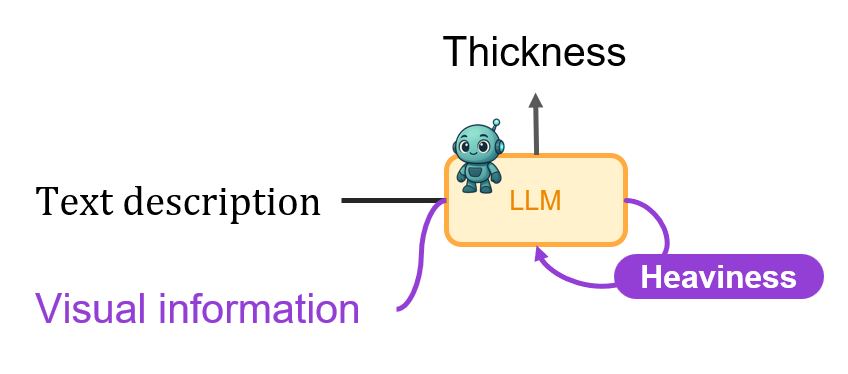}
    
    \caption{Ours}
    \label{fig:short-b}
  \end{subfigure}
  \caption{\textbf{Heaviness-aware thickness refinement (HAT).} (a) NeRF2Physics assigns thickness based on material priors. (b) Our HAT module first classifies the object as \emph{heavy} or \emph{light} and refines thickness prediction toward physically plausible ranges}
  \label{fig:short}
\end{figure}

\noindent\textbf{Heaviness-aware thickness.} We found that the material prediction has weak correlation to mass (proof in Supplementary), which is surprising given the expected value of the mass is the multiplication of mass density and volume. This motivated us to focus on improving thickness prediction. According to \cref{tab:light_heavy}, the current thickness method is performing worse on heavy objects. Therefore, we propose a thickness estimation method that are more mass-aware, using GPT4 to refine the mass based on heaviness of the object. We first use SiPhy's initial predictions to classify the object as \emph{heavy} or \emph{light}. This classification is then used to adjust thickness prediction, biasing the prediction toward thickness ranges that are physically plausible (\cref{fig:short}).


    


\section{Experiments}
\label{sec:experiment}


\noindent\textbf{Dataset.}
We evaluate on three benchmarks. \textbf{ABO-500}~\cite{DBLP:journals/corr/abs-2110-06199} contains 500 everyday objects with ground-truth mass; we follow the standard 300/100/100 train/val/test split and use a single RGB view per object. \textbf{MVImgNet-100}~\cite{yu2023mvimgnet} is a large-scale real-world multi-view dataset; we use 100 randomly sampled objects with a single front-facing view as input. \textbf{PhysXNet-100}~\cite{cao2025physx} is a synthetic dataset with dense per-pixel physical property annotations including mass density and Young's modulus; we randomly sample 100 scenes for evaluation.

\noindent\textbf{Baselines.} \textbf{NeRF2Physics}~\cite{zhai2024physical} uses NeRF to reconstruct the object, fuses CLIP features at each 3D point, and estimates physical properties via retrieval-based aggregation before volumetric mass integration. \textbf{PUGS}~\cite{shuai2025pugszeroshotphysicalunderstanding} adopts 3D Gaussian Splatting with shape- and region-aware contrastive losses, propagating CLIP features across Gaussians and computing mass from per-Gaussian density and volume. \textbf{GaussianProperty}~\cite{xu2024gaussianproperty} lifts CLIP features onto Gaussian primitives for property estimation. \textbf{LLaVA}~\cite{liu2023visualinstructiontuning} predicts mass directly from a single image using a VLM, following the evaluation protocol of NeRF2Physics.

\definecolor{lightblue}{RGB}{220,235,247}

\begin{table}[ht!]
  \centering
  \scriptsize
  \tabcolsep=0.01in
    \caption{Mass prediction evaluation on the ABO-500 test set. ``M'' and ``S'' denote multi-view and single-view inputs.}
  \resizebox{0.96\linewidth}{!}{%
    \begin{tabular}{lccccc}
      \toprule
      Method & View & ADE ($\downarrow$) & ALDE ($\downarrow$) & APE ($\downarrow$) & MnRE ($\uparrow$) \\
      \midrule
      NeRF2Physics~\cite{zhai2024physical} & M & 8.74 & 0.78 & 1.06 & 0.55 \\
      PUGS~\cite{shuai2025pugszeroshotphysicalunderstanding} & M & 30.30 & 1.59 & 7.68 & 0.30 \\

      LLaVA~\cite{liu2023visualinstructiontuning} & S & 17.33 & 1.89 & 1.84 & 0.31 \\

      Image2Mass~\cite{pmlr-v78-standley17a} & S & 12.50 & 1.79 & \textbf{0.98} & 0.31 \\ 
    
      SiPhy (uniform 2D sampling) & S & 11.94 & 1.27 & 2.61 & 0.39 \\

      SiPhy (average SAM masks) & S & 10.23 & 0.77 & 1.00 & 0.54 \\
      \midrule

       SiPhy (Ours) & S & \textbf{7.78} & \textbf{0.74} & 1.00 & \textbf{0.58} \\
      \bottomrule
    \end{tabular}%
  }

  \label{tab:mass}
\end{table}
\begin{table}[t!]
\centering
\footnotesize

\begin{minipage}[t]{0.55\textwidth}
\centering
\setlength{\tabcolsep}{4pt}
\renewcommand{\arraystretch}{0.95}
\caption{Material segmentation performance across datasets.}
\label{tab:material_miou}
\resizebox{\textwidth}{!}{
\begin{tabular}{llcc}
\toprule
\textbf{Dataset} & \textbf{Model} & \textbf{mIoU ($\uparrow$)} & \textbf{M-mIoU ($\uparrow$)} \\
\midrule
 \multirow{4}{*}{ABO500~\cite{DBLP:journals/corr/abs-2110-06199}}
      & NeRF2Physics~\cite{zhai2024physical}      & 0.18 & 0.31 \\
      & PUGS~\cite{shuai2025pugszeroshotphysicalunderstanding}               & 0.22 & 0.40 \\
      & GaussianProperty~\cite{xu2024gaussianproperty} & \textbf{0.29} & \textbf{0.49} \\ \cmidrule{2-4}
      & SiPhy (Ours)               & 0.25 & 0.42 \\
    \midrule
    \multirow{4}{*}{MVImgNet100~\cite{yu2023mvimgnet}}
      & NeRF2Physics~\cite{zhai2024physical}       & - & - \\
      & PUGS~\cite{shuai2025pugszeroshotphysicalunderstanding}               & - & - \\
      & GaussianProperty~\cite{xu2024gaussianproperty} & \textbf{0.19} & 0.23 \\ \cmidrule{2-4}
      & SiPhy (Ours)               & \textbf{0.19} & \textbf{0.25} \\
    \midrule
    \multirow{4}{*}{PhysXNet100~\cite{cao2025physx}}
      & NeRF2Physics~\cite{zhai2024physical}       & 0.03 & 0.06 \\
      & PUGS~\cite{shuai2025pugszeroshotphysicalunderstanding}              & 0.04 & 0.07 \\
      & GaussianProperty~\cite{xu2024gaussianproperty} & \textbf{0.14} & 0.21 \\ \cmidrule{2-4}
      & SiPhy (Ours)              & \textbf{0.14} & \textbf{0.31} \\
\bottomrule
\end{tabular}
}
\end{minipage}
\hfill
\begin{minipage}[t]{0.43\textwidth}
\centering
\setlength{\tabcolsep}{6pt}
\renewcommand{\arraystretch}{1.05}
\caption{Material density and Young's Modulus estimation on PhysXNet100.}
\label{tab:other_properties}
\resizebox{\textwidth}{!}{
\begin{tabular}{l l c c}
\toprule
\textbf{Property} & \textbf{Model} & \textbf{View} & \textbf{MAE} ($\downarrow$) \\
\midrule
      \multirow{3}{*}{\shortstack{Density\\(kg/m$^3$)}} 
       & NeRF2Physics & M  & 2044 \\
      & PUGS        & M  & \textbf{1297} \\ 
      \cmidrule{2-4}
      & SiPhy (Ours)         & S & 1315 \\ \midrule
      \multirow{3}{*}{\shortstack{Young \\ Modulus \\ (GPa)}} 
       & NeRF2Physics & M  & 68 \\
      & PUGS        & M  & 68 \\
      \cmidrule{2-4}
      & SiPhy (Ours)         & S & \textbf{52} \\
      
\bottomrule
\label{tab:material_segmentation}
\end{tabular}
}
\end{minipage}

\end{table}

\noindent\textbf{Implementation.} For part-based contrastive alignment, we apply a local self-attention module restricted to points sharing the same SAM-derived part. Each CLIP image embedding ($D{=}512$) is normalized and passed through an attention layer with attention dimension $64$. We optimize the supervised contrastive loss using Adam with a learning rate of $1\times 10^{-4}$ and temperature $\tau=0.1$.

For vision–language feature fusion, we use OpenCLIP ViT-B/16 pretrained on 
DataComp-1B. LLM responses for material attributes, including density and Young’s modulus, 
are generated using GPT-4. We set the number of candidate materials to $K=5$ and sampling temperature to $T=0.1$. An ablation on $K \in \{3, 5, 10\}$ is provided in the Supplementary. Captions are generated using Instructional BLIP-2~\cite{instructblip} with Flan-T5-XL.


\subsection{Quantitative Comparison to SOTA methods}

\begin{figure*}[t]
  \centering
  \includegraphics[width=\linewidth]{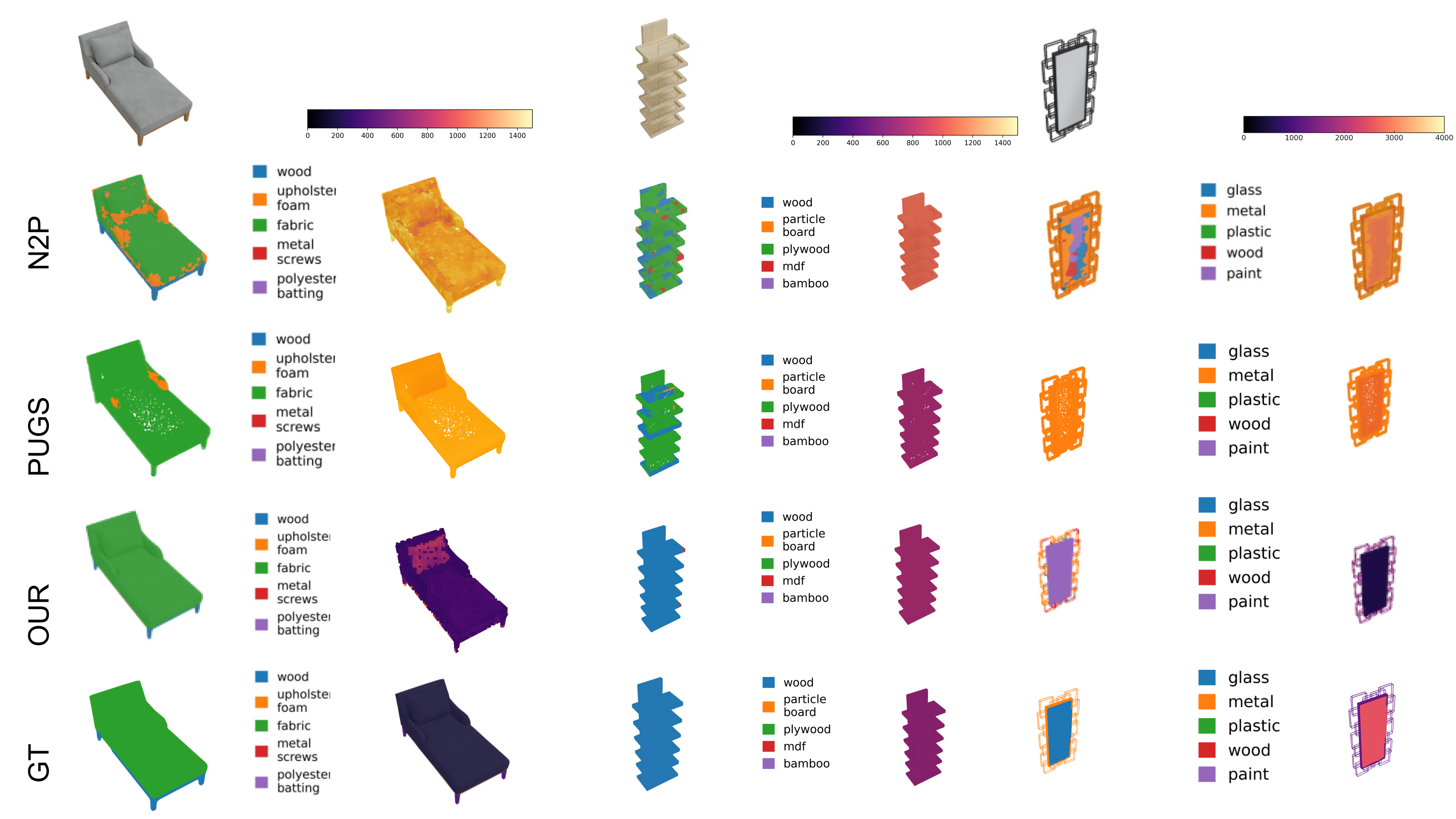}
  \caption{\textbf{Qualitative results on ABO500. Material segmentation (left) and mass density (right).} SiPhy yields more coherent material segmentation than NeRF2Physics and PUGS, especially on thin structures and fine-grained parts. }
  \label{fig:abo_qualitative}
\end{figure*}

\begin{figure*}[t]
  \centering
  \includegraphics[width=\linewidth]{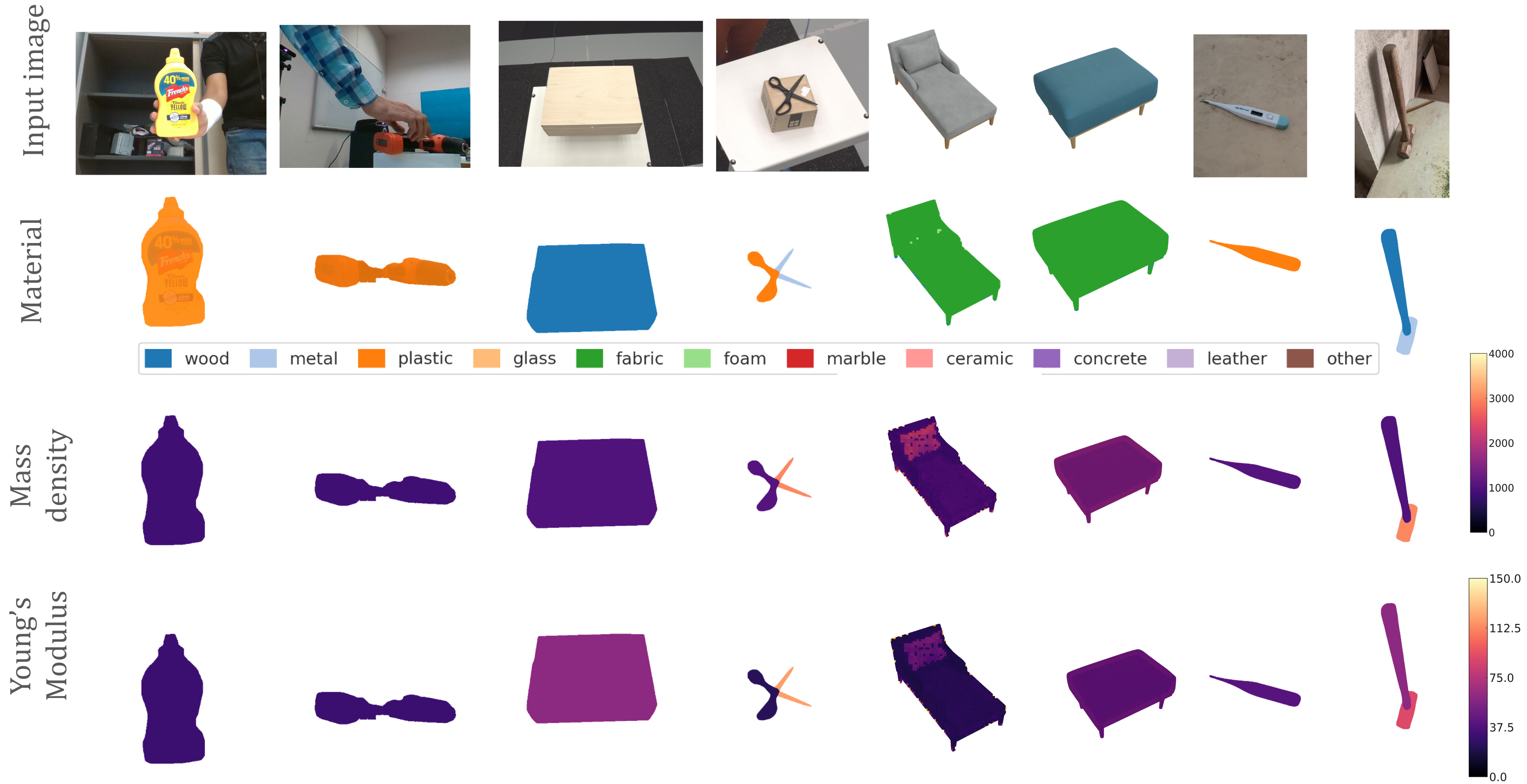}
  \caption{\textbf{Qualitative results on HO3D, ARCTIC, ABO500, and MVImgNET.} From a single image, SiPhy predicts material labels, mass density, and Young’s modulus with consistent and physically plausible outputs across real and synthetic settings. All property maps use a jet colormap; mass density ranges 0--3000~kg/m$^3$ and Young's modulus 0--100~GPa. Objects~5 and~6 have markedly lower values and use narrower ranges (density 0--1500~kg/m$^3$, Young's modulus 0--20~GPa).}
  \label{fig:ho3d_qualitative}
\end{figure*}


\noindent\textbf{Mass Evaluation.} We evaluate mass prediction using the same metrics as NeRF2Physics~\cite{zhai2024physical}:  
Absolute Difference Error (ADE), Absolute Log Difference Error (ALDE), Absolute Percentage Error (APE), and Minimum Ratio Error (MnRE). MnRE is considered the most reliable metric due to its scale invariance. To ensure a fair and comprehensive comparison, We report results on both the official test split and the full dataset for all models. GaussianProperty~\cite{xu2024gaussianproperty} does not release mass-estimation code and cannot be evaluated.

As shown in Table~\ref{tab:mass}, \methodname consistently outperforms all baselines across major metrics. On the primary metric MnRE, our method improves over NeRF2Physics by \textbf{5.5\%} and over PUGS by \textbf{93.3\%}, demonstrating strong robustness despite relying only on a single-view input. Our ADE is also substantially lower (7.78 vs. 8.74 for NeRF2Physics and 30.30 for PUGS), indicating improved absolute mass accuracy. These results highlight the advantage of single-view reasoning with structured vision–language physical inference.

\noindent\textbf{Material Segmentation Evaluation.}
As shown in Table~\ref{tab:material_miou}, SiPhy achieves strong material segmentation across all three datasets. On ABO-500, SiPhy improves over NeRF2Physics by \textbf{38.9\%} mIoU and over PUGS by \textbf{13.6\%} mIoU; GaussianProperty scores higher due to its multi-view input. On MVImgNet-100, SiPhy matches GaussianProperty in mIoU and improves M-mIoU by \textbf{8.7\%}. On PhysXNet-100, SiPhy achieves the largest gains, outperforming all baselines including GaussianProperty (\textbf{+47.6\%} M-mIoU).





\noindent\textbf{Density and Young’s Modulus Evaluation.} We further evaluate density and Young’s modulus prediction on the PhysXNet100 dataset, which provides per-pixel annotations for both properties. Following prior work, we compute mean absolute error (MAE) between the predicted and ground-truth maps. As shown in Table~\ref{tab:other_properties}, \methodname achieves competitive or superior performance compared to multi-view baselines. On density estimation, SiPhy reduces the MAE by \textbf{35.5\%} compared to NeRF2Physics, despite using only a single-view input. SiPhy performs comparably to PUGS, with only a small \(1.4\%\) gap, even though PUGS relies on full multi-view Gaussian reconstruction. For Young’s modulus, SiPhy achieves the best performance, reducing MAE by \textbf{23.5\%} relative to both NeRF2Physics and PUGS. These results demonstrate that our vision–language physical reasoning and part-based alignment generalize effectively to fine-grained material attributes, outperforming reconstruction-heavy multi-view methods under the single-view setting.





\subsection{Ablation Study}


\begin{table}[t!]
\centering
\footnotesize

\begin{minipage}[t]{0.55\textwidth}
  \centering
  \footnotesize
  \setlength{\tabcolsep}{4pt}
  \renewcommand{\arraystretch}{1.05}
  \caption{Performance comparison between Our method and NeRF2Physics on ABO500 splits (heavy and light objects), and the effect of the \textbf{heaviness-aware thickness (\emph{HAT})} module. Improvements are marked in \textcolor{dpurple}{purple}.}
  \label{tab:light_heavy}
  \resizebox{\columnwidth}{!}{
  \begin{tabular}{llcccc}
    \toprule
    Model & Split & ADE ($\downarrow$) & ALDE ($\downarrow$) & APE ($\downarrow$) & MnRE ($\uparrow$) \\
    \midrule
    \raisebox{-1.25\normalbaselineskip}[0pt][0pt]{\parbox{3.2em}{Ours\\[-1pt]\emph{w/o HAT}}}
        & ABO500       & 9.84  & 0.81 & 0.85 & 0.53 \\
        & ABO500-heavy & 18.86 & 0.64 & 0.54 & 0.57 \\
        & ABO500-light & 2.07  & 0.73 & 0.57 & 0.58 \\
    \midrule
    \raisebox{-1.25\normalbaselineskip}[0pt][0pt]{Ours}
        & ABO500
        & 7.78\({}^{\mkern-8mu\mbox{\tiny\myhat{+2.06}}}\)
        & 0.74\({}^{\mkern-8mu\mbox{\tiny\myhat{+0.07}}}\)
        & 1.00\({}^{\mkern-8mu\mbox{\tiny\myhat{-0.15}}}\)
        & 0.58\({}^{\mkern-8mu\mbox{\tiny\myhat{+0.05}}}\) \\
        & ABO500-heavy
        & 15.59\({}^{\mkern-8mu\mbox{\tiny\myhat{+3.27}}}\)
        & 0.48\({}^{\mkern-8mu\mbox{\tiny\myhat{+0.16}}}\)
        & 0.46\({}^{\mkern-8mu\mbox{\tiny\myhat{+0.08}}}\)
        & 0.65\({}^{\mkern-8mu\mbox{\tiny\myhat{+0.08}}}\) \\
        & ABO500-light
        & 3.57\({}^{\mkern-8mu\mbox{\tiny\myhat{-1.50}}}\)
        & 0.85\({}^{\mkern-8mu\mbox{\tiny\myhat{-0.12}}}\)
        & 1.29\({}^{\mkern-8mu\mbox{\tiny\myhat{-0.72}}}\)
        & 0.54\({}^{\mkern-8mu\mbox{\tiny\myhat{-0.04}}}\) \\
    \bottomrule
  \end{tabular}
  }
\end{minipage}
\hfill
\begin{minipage}[t]{0.43\textwidth}
\centering
\caption{Sensitivity analysis under independent perturbations of estimated depth and camera intrinsics on ABO-500.}
\vspace{-0.3cm}
\label{tab:sensitivity}
\resizebox{\columnwidth}{!}{%
\begin{tabular}{lcccc}
\hline
Setting & ADE ($\downarrow$) & ALDE ($\downarrow$) & APE ($\downarrow$) & MnRE ($\uparrow$) \\
\hline
Depth $\times$ 0.9  & 8.08 & 0.77 & 0.87 & \textbf{0.56} \\
Depth $\times$ 1.1  & 9.14 & 0.75 & 1.27 & \textbf{0.56} \\
\hline
Intrinsics $\times$ 0.9 & 9.17 & 0.75 & 1.28 & \textbf{0.56} \\
Intrinsics $\times$ 1.1 & 8.07 & 0.76 & 0.87 & \textbf{0.57} \\
\hline
\end{tabular}
}
\end{minipage}

\end{table}

\noindent\textbf{Effectiveness of heaviness-aware thickness refinement.}
Table~\ref{tab:light_heavy} compares SiPhy with and without HAT on the ABO-500 heavy/light splits. Without HAT, mass prediction is substantially worse for heavy objects, with ADE increasing from 2.07 on light objects to 18.86 on heavy objects. HAT improves heavy-object ADE from 18.86 to 15.59 (\textbf{17.3\%} improvement) and MnRE from 0.57 to 0.65, leading to better overall ABO-500 performance. This suggests that heaviness-aware thickness refinement mainly benefits dense or bulky objects, where volume errors have a larger impact on mass estimation.


Sensitivity of the spacing formula to depth and intrinsics estimation errors is analyzed in the Supplementary materials.

\begin{figure}[ht!]
\vspace{-0.25cm}
  \centering
  \begin{subfigure}[t]{0.49\linewidth}
    \centering
    \includegraphics[width=\linewidth]{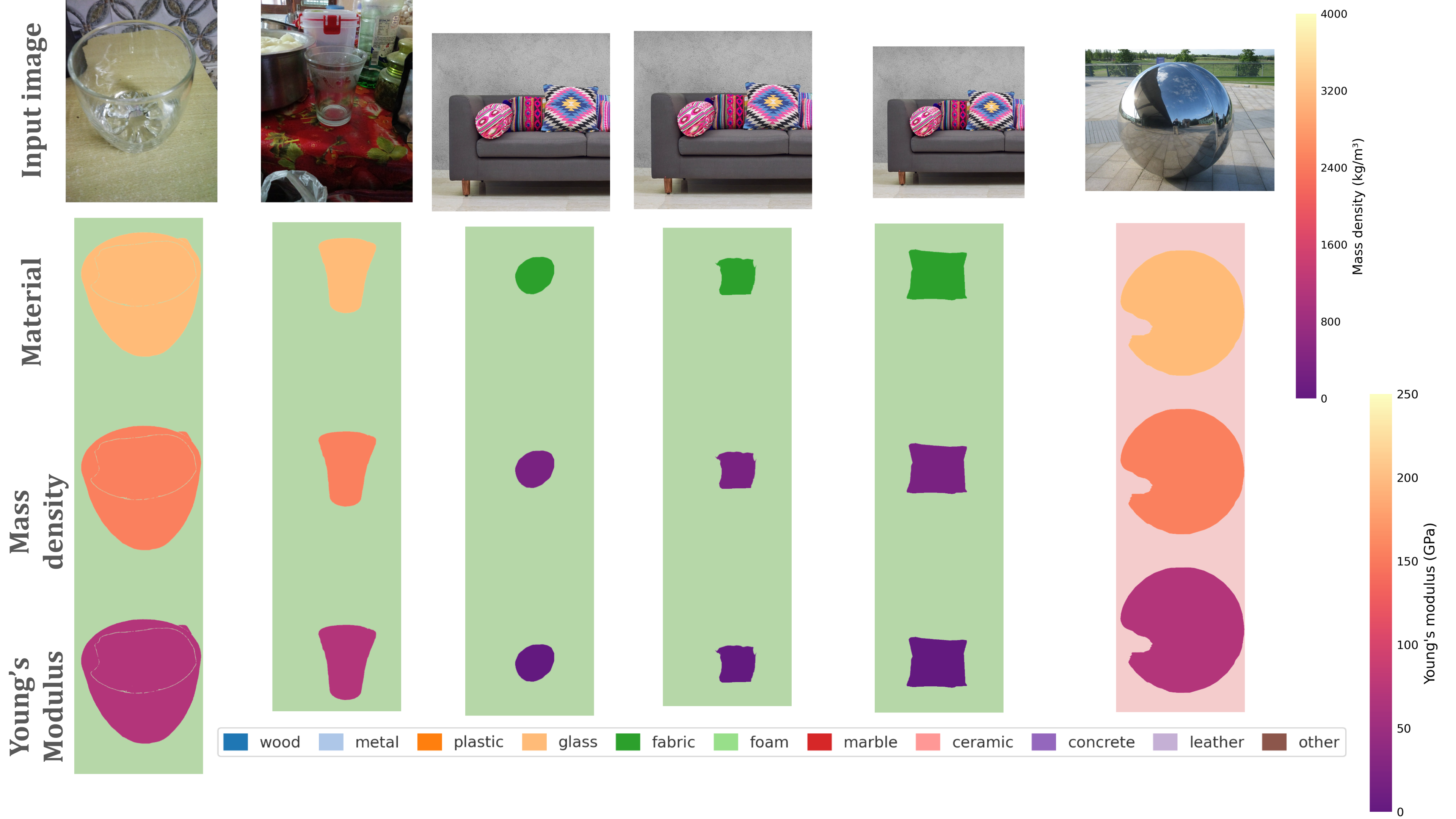}
    \caption{SiPhy}
    \label{fig:edge_objects}
  \end{subfigure}
  \hfill
  \begin{subfigure}[t]{0.49\linewidth}
    \centering
    \includegraphics[width=\linewidth]{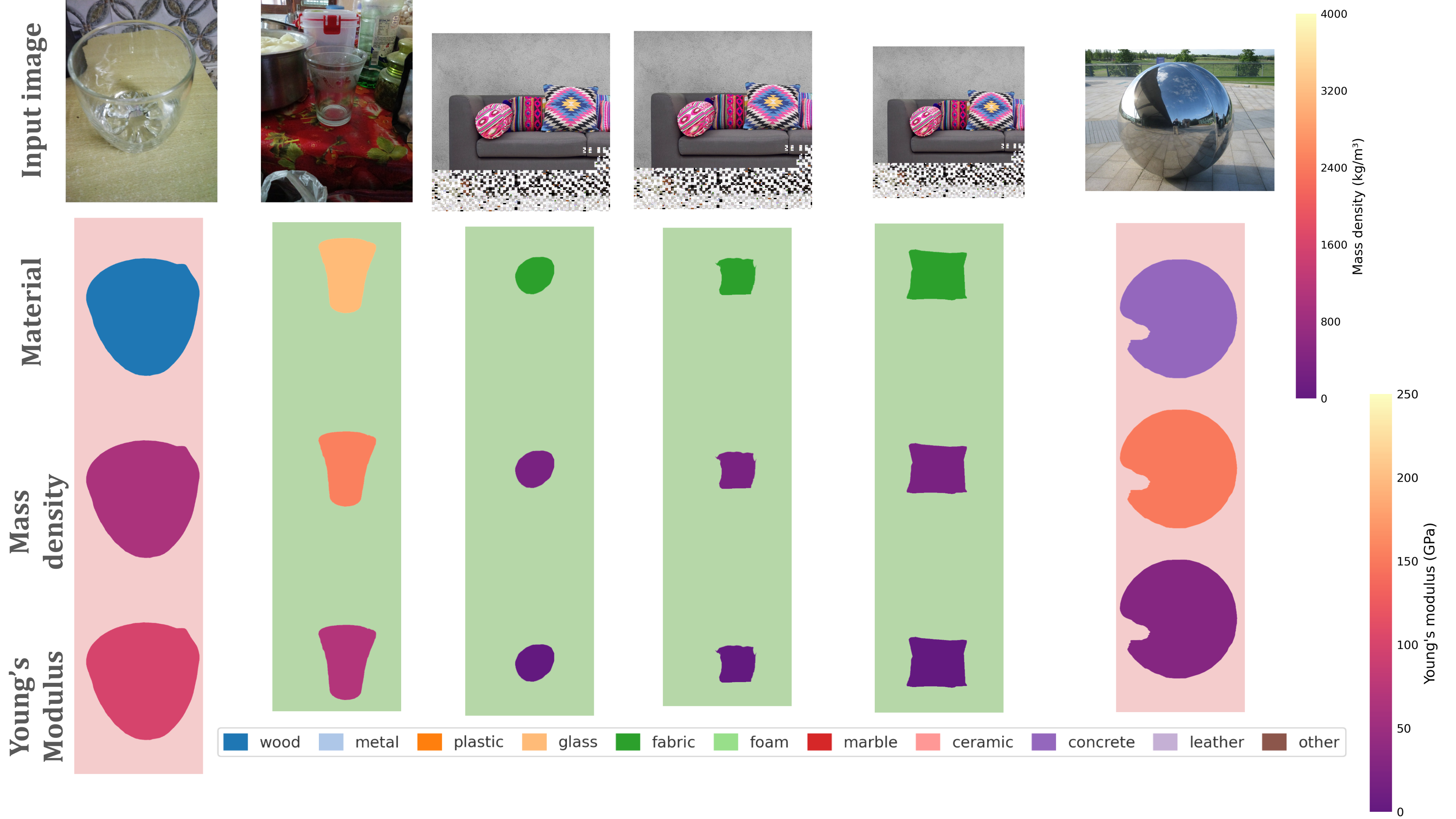}
    \caption{LLaVA}
    \label{fig:edge_objects_llava}
  \end{subfigure}
  \vspace{-0.35cm}
  \caption{\textbf{Edge-case analysis.} From left to right: the first 2 objects are \textbf{transparent}, the next 3 (pillows, same image) are \textbf{highly textured}, and the last is \textbf{reflective}. Predictions are bordered with a \textcolor{green!60!black}{green} box if correct and a \textcolor{red}{red} box if incorrect. SiPhy handles transparent and textured objects well; reflective surfaces remain a failure mode for both methods.}
  \label{fig:edge_objects_compare}
  \vspace{-0.3cm}
\end{figure}

\noindent\textbf{Edge-case analysis.}
We test SiPhy on transparent, highly textured, and reflective objects (\cref{fig:edge_objects_compare}). SiPhy handles most transparent and textured cases, while reflective surfaces remain challenging because specular highlights can mislead both SAM segmentation and VLM material prediction.


\subsection{Real-world Application}
We evaluate two downstream uses of SiPhy. For hand-object interaction, SiPhy can annotate physical properties for datasets that lack such labels. For image-to-audio generation, SiPhy provides material priors for sound synthesis, where material is often more informative than appearance alone.


\noindent\textbf{Hand-object Interaction.}
HO3D~\cite{hampali2020honnotate} and ARCTIC~\cite{fan2023arctic} are real-world hand-object interaction datasets capturing everyday objects featuring partial hand occlusion with varied illumination. We evaluate on 9 objects from HO3D and 10 from ARCTIC, selecting a single front-view RGB frame per object. Ground-truth masses are sourced from manufacturer product specifications. Since multi-view baselines (NeRF2Physics, PUGS) require multi-view captures unavailable in these datasets, we compare against LLaVA as the strongest applicable single-view baseline.
As shown in Table~\ref{tab:real_world_mass} and Table~\ref{tab:real_world_other}, SiPhy outperforms LLaVA in mass prediction (MnRE 0.67 vs.\ 0.58 on HO3D; 0.57 vs.\ 0.53 on ARCTIC) and achieves lower error in density and Young’s modulus, confirming its effectiveness as a physical-property annotator for real-world interaction data (\cref{fig:ho3d_qualitative}).

\begin{table}[t!]
\centering
\footnotesize

\begin{minipage}[t]{0.55\textwidth}
\centering
\setlength{\tabcolsep}{2pt}
\renewcommand{\arraystretch}{1}
\caption{Material segmentation performance across datasets.}
\label{tab:real_world_mass}
\resizebox{\textwidth}{!}{
\begin{tabular}{llcccc}
\toprule
Dataset & Model & ADE($\downarrow$) & ALDE($\downarrow$) & APE($\downarrow$) & MnRE($\uparrow$) \\
\midrule

{HO3D~\cite{hampali2020honnotate}}
& LLaVA~\cite{liu2023visualinstructiontuning}& 0.24 & \textbf{0.48} & 0.58 & 0.58 \\
& SiPhy (Ours) & \textbf{0.19} & 0.50 & \textbf{0.41} & \textbf{0.67} \\
\midrule

{ARCTIC~\cite{fan2023arctic}}
& LLaVA~\cite{liu2023visualinstructiontuning} & \textbf{0.21} & \textbf{0.60} & 0.50 & 0.53 \\
& SiPhy (Ours) & 0.26 & 0.92 & \textbf{0.46} & \textbf{0.57} \\
\bottomrule
\end{tabular}
}
\end{minipage}
\hfill
\begin{minipage}[t]{0.43\textwidth}
\centering
\setlength{\tabcolsep}{6pt}
\renewcommand{\arraystretch}{1.05}
\caption{Material density and Young's Modulus estimation on PhysXNet100.}
\label{tab:real_world_other}
\resizebox{\textwidth}{!}{
\begin{tabular}{llcc}
\toprule
Dataset & Model & Property & ME ($\downarrow$) \\
\midrule

{HO3D~\cite{hampali2020honnotate}}
& LLaVA~\cite{liu2023visualinstructiontuning} & Mass density & 1308 \\
& SiPhy (Ours) & Mass density & \textbf{1192} \\
\midrule

{ARCTIC~\cite{fan2023arctic}}
& LLaVA~\cite{liu2023visualinstructiontuning} & Young's modulus & 74 \\
& SiPhy (Ours) & Young's modulus & \textbf{62} \\
\bottomrule

\end{tabular}
}
\end{minipage}

\end{table}

\begin{figure}[t]
  \centering
  \includegraphics[width=\linewidth]{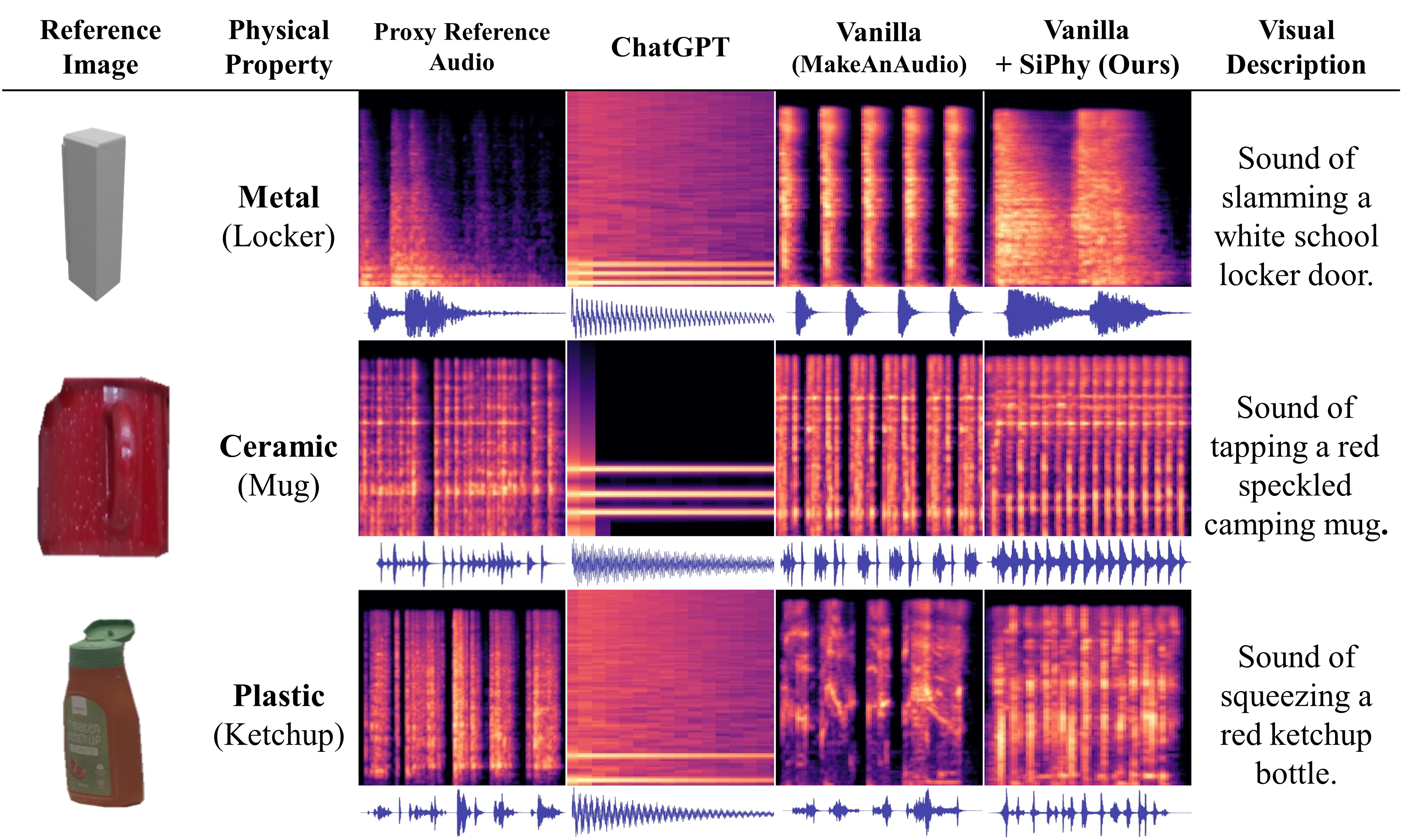}
  \caption{\textbf{Material-grounded audio generation.} Mel-spectrograms for three objects under ChatGPT-only, Vanilla (VLM caption $\to$ MakeAnAudio), and Vanilla + SiPhy. SiPhy's material prior corrects appearance-driven errors (e.g., metal locker misidentified as wood), producing spectrograms that better match the reference audio.}
  \label{fig:audio_main}
\end{figure}

\noindent\textbf{Image-to-Audio Generation.}
\label{subsec:audio} 
Object acoustics depend on material, not appearance. When visual texture misleads, appearance-driven systems produce implausible audio; SiPhy's material prior corrects this bias. 
We compare ChatGPT-only, Vanilla (VLM caption $\to$ MakeAnAudio~\cite{huang2023make}), and Vanilla + SiPhy. As shown in \cref{fig:audio_main}, SiPhy's material prior corrects appearance-driven errors (e.g., a metal locker misidentified as wood), producing spectrograms that better match the reference. Full results are in the Supplementary.

\section{Conclusion}
We presented SiPhy, a single-image framework that unifies geometric cues,
semantic understanding, and language-driven physical knowledge to estimate both pixel-level and object-level physical properties. Despite operating without multi-view
supervision, SiPhy achieves strong performance across synthetic and real-world
datasets, enabled by part-based contrastive alignment and heaviness-aware
refinement for volume estimation. SiPhy offers a scalable alternative to reconstruction-heavy pipelines, but
remains constrained by the quality of single-view geometric priors and the
inherent ambiguity of inferring physical attributes from one RGB image.
Transparent, reflective, and highly textured objects remain challenging because depth estimation and CLIP-based material recognition can become unreliable. HAT currently uses a discrete heavy/light prior; extending it to continuous or compositional thickness estimation is left for future work.

\section*{Acknowledgment}
This work was supported by Michigan State University. We are grateful to Xiaoming Liu for an insightful discussion that inspired our experiment on generating sounds from material properties.

\bibliographystyle{splncs04}
\bibliography{main}
\end{document}